\documentclass[letterpaper, 10 pt, conference]{class/ieeeconf}  % Comment this line out
\IEEEoverridecommandlockouts                              % This command is only
\usepackage{xcolor}
\usepackage{siunitx}  
\usepackage{graphicx}
 \graphicspath{./figures/robot_demonstration}
\usepackage{amsmath}
\usepackage{amssymb}
\usepackage{latexsym}
\usepackage{url}
\usepackage{cite}
\usepackage{relsize}
\usepackage{multirow}
\usepackage{afterpage}
\usepackage{ifthen}
\usepackage{graphicx}
\usepackage{algpseudocode,algorithm,algorithmicx}
\usepackage{subfigure}
\usepackage{epstopdf}
\usepackage{stackengine}
\usepackage{textcomp} % new

\usepackage{gensymb}
\usepackage{booktabs}
\usepackage{makecell}
\usepackage{hhline}
\usepackage{courier}
\usepackage{lipsum}
\usepackage{pifont}
\usepackage{bm}
\usepackage{flushend}
\usepackage{xcolor} % new
\usepackage{times}
\usepackage{epsfig}
\usepackage{graphicx}
\usepackage{amsmath}
\usepackage{amssymb}
\usepackage{pifont}
\usepackage{color}
\usepackage{multirow}
\usepackage{overpic}
\usepackage{arydshln}
\usepackage{makecell}
\usepackage{times}
\usepackage[T1]{fontenc}
\usepackage{hhline}
\usepackage{xcolor}
\usepackage{pifont}
\usepackage{colortbl}
\usepackage{verbatim}
\usepackage{bm}
\usepackage{wrapfig}
\usepackage{booktabs}
\usepackage{eso-pic}
\usepackage{placeins}

\definecolor{darkpastelgreen}{rgb}{0.01, 0.75, 0.24}
\definecolor{darkpink}{rgb}{0.91, 0.33, 0.5}
\definecolor{mygray}{gray}{.94}

\definecolor{linkcolor}{RGB}{255,0,0}
\definecolor{urlcolor}{RGB}{255,105,180}
\definecolor{citecolor}{RGB}{0, 80, 200}
\definecolor{citecolor1}{RGB}{0,153,255}

\makeatletter
\let\NAT@parse\undefined
\makeatother
\usepackage[bookmarks=false, linkcolor=blue, urlcolor=blue, citecolor=blue]{hyperref} 

\hypersetup{
    colorlinks=true,
    linkcolor=red,
    filecolor=magenta,      
    urlcolor=blue,
    pdfstartview={FitH},
    citecolor =blue
    }
    
\definecolor{mygray}{gray}{.94}

\usepackage{graphicx}

\usepackage{xspace}

\providecommand{\kuka}{\textsc{KUKA} LBR iiwa R820\xspace}

\newcommand{\etal}{\textit{et al.}}

\title{\LARGE \bf Open-Vocabulary Affordance Detection using Knowledge Distillation and Text-Point Correlation}

\author{Tuan Van Vo$^{1}$, Minh Nhat Vu$^{2}$, Baoru Huang$^3$, Toan Nguyen$^{1}$, Ngan Le$^4$, Thieu Vo$^5$, Anh Nguyen$^6$
\thanks{$^1$ FPT Software AI Center, Vietnam {\tt tuanvv7@fpt.com}}
\thanks{$^2$ Automation \& Control Institute, TU Wien, Vienna, Austria %{\tt vu@acin.tuwien.ac.at}
}
\thanks{$^3$ Imperial College London, UK
}
\thanks{$^4$ Department of Computer Science, University of Arkansas, USA 
}
\thanks{$^5$ Faculty of Mathematics and Statistics, Ton Duc Thang University, Ho Chi Minh city, Vietnam} %{\tt vongocthieu@tdtu.edu.vn}}
\thanks{$^5$ Department of Computer Science, University of Liverpool, UK %{\tt anh.nguyen@liverpool.ac.uk}
}}

\begin{document}
% Macros

\newtheorem{problem}{Problem}
\newtheorem{lemma}{Lemma}
\newtheorem{theorem}[lemma]{Theorem}
\newtheorem{claim}{Claim}
\newtheorem{corollary}[lemma]{Corollary}
\newtheorem{definition}[lemma]{Definition}
\newtheorem{proposition}[lemma]{Proposition}
\newtheorem{remark}[lemma]{Remark}
\newenvironment{LabeledProof}[1]{\noindent{\it Proof of #1: }}{\qed}

\def\beq#1\eeq{\begin{equation}#1\end{equation}}
\def\bea#1\eea{\begin{align}#1\end{align}}
\def\beg#1\eeg{\begin{gather}#1\end{gather}}
\def\beqs#1\eeqs{\begin{equation*}#1\end{equation*}}
\def\beas#1\eeas{\begin{align*}#1\end{align*}}
\def\begs#1\eegs{\begin{gather*}#1\end{gather*}}

\newcommand{\poly}{\mathrm{poly}}
\newcommand{\eps}{\epsilon}
\newcommand{\e}{\epsilon}
\newcommand{\polylog}{\mathrm{polylog}}
\newcommand{\rob}[1]{\left( #1 \right)} %Round Brackets
\newcommand{\sqb}[1]{\left[ #1 \right]} %square Brackets
\newcommand{\cub}[1]{\left\{ #1 \right\} } %curly brackets
\newcommand{\rb}[1]{\left( #1 \right)} %Round
\newcommand{\abs}[1]{\left| #1 \right|} %| |
\newcommand{\zo}{\{0, 1\}}
\newcommand{\zonzo}{\zo^n \to \zo}
\newcommand{\zokzo}{\zo^k \to \zo}
\newcommand{\zot}{\{0,1,2\}}
\newcommand{\en}[1]{\marginpar{\textbf{#1}}}
\newcommand{\efn}[1]{\footnote{\textbf{#1}}}
\newcommand{\vecbm}[1]{\boldmath{#1}} %more general (handles greek letters)
\newcommand{\uvec}[1]{\hat{\vec{#1}}}
\newcommand{\thv}{\vecbm{\theta}}
\newcommand{\junk}[1]{}
\newcommand{\var}{\mathop{\mathrm{var}}}
\newcommand{\rank}{\mathop{\mathrm{rank}}}
\newcommand{\diag}{\mathop{\mathrm{diag}}}
\newcommand{\tr}{\mathop{\mathrm{tr}}}
\newcommand{\acos}{\mathop{\mathrm{acos}}}
\newcommand{\atantwo}{\mathop{\mathrm{atan2}}}
\newcommand{\SVD}{\mathop{\mathrm{SVD}}}
\newcommand{\quadf}{\mathop{\mathrm{q}}}
\newcommand{\linterp}{\mathop{\mathrm{l}}}
\newcommand{\sgn}{\mathop{\mathrm{sign}}}
\newcommand{\sym}{\mathop{\mathrm{sym}}}
\newcommand{\avg}{\mathop{\mathrm{avg}}}
\newcommand{\mean}{\mathop{\mathrm{mean}}}
\newcommand{\erf}{\mathop{\mathrm{erf}}}
\newcommand{\grad}{\nabla}
\newcommand{\R}{\mathbb{R}}
\newcommand{\defeq}{\triangleq}
\newcommand{\dims}[2]{[#1\!\times\!#2]}
\newcommand{\sdims}[2]{\mathsmaller{#1\!\times\!#2}}
\newcommand{\udims}[3]{#1}
\newcommand{\udimst}[4]{#1}
\newcommand{\com}[1]{\rhd\text{\emph{#1}}}
\newcommand{\ind}{\hspace{1em}}
\newcommand{\argmin}[1]{\underset{#1}{\operatorname{argmin}}}
\newcommand{\floor}[1]{\left\lfloor{#1}\right\rfloor}
\newcommand{\step}[1]{\vspace{0.5em}\noindent{#1}}
\newcommand{\quat}[1]{\ensuremath{\mathring{\mathbf{#1}}}}
\newcommand{\norm}[1]{\left\lVert#1\right\rVert}
\newcommand{\ignore}[1]{}
\newcommand{\specialcell}[2][c]{\begin{tabular}[#1]{@{}c@{}}#2\end{tabular}}
\newcommand*\Let[2]{\State #1 $\gets$ #2}
\newcommand{\algorithmicbreak}{\textbf{break}}
\newcommand{\Break}{\State \algorithmicbreak}
\newcommand{\ra}[1]{\renewcommand{\arraystretch}{#1}}

\renewcommand{\vec}[1]{\mathbf{#1}} %looks better

\algdef{S}[FOR]{ForEach}[1]{\algorithmicforeach\ #1\ \algorithmicdo}
\algnewcommand\algorithmicforeach{\textbf{for each}}
\algrenewcommand\algorithmicrequire{\textbf{Require:}}
\algrenewcommand\algorithmicensure{\textbf{Ensure:}}
\algnewcommand\algorithmicinput{\textbf{Input:}}
\algnewcommand\INPUT{\item[\algorithmicinput]}
\algnewcommand\algorithmicoutput{\textbf{Output:}}
\algnewcommand\OUTPUT{\item[\algorithmicoutput]}

\maketitle

\thispagestyle{empty}
\pagestyle{empty}

%%%%%%%%%%%%%%%%%%%%%%%%%%%%%%%%%%%%%%%%%%%%%%%%%%%%%%%%%%%%%%%%%%%%%%%%%%%%%%%%
\begin{abstract}
Affordance detection presents intricate challenges and has a wide range of robotic applications. Previous works have faced limitations such as the complexities of 3D object shapes, the wide range of potential affordances on real-world objects, and the lack of open-vocabulary support for affordance understanding. In this paper, we introduce a new open-vocabulary affordance detection method in 3D point clouds, leveraging knowledge distillation and text-point correlation. Our approach employs pre-trained 3D models through knowledge distillation to enhance feature extraction and semantic understanding in 3D point clouds. We further introduce a new text-point correlation method to learn the semantic links between point cloud features and open-vocabulary labels. The intensive experiments show that our approach outperforms previous works and adapts to new affordance labels and unseen objects. Notably, our method achieves the improvement of $7.96\%$ mIOU score compared to the baselines. Furthermore, it offers real-time inference which is well-suitable for robotic manipulation applications.

\end{abstract}

%%%%%%%%%%%%%%%%%%%%%%%%%%%%%%%%%%%%%%%%%%%%%%%%%%%%%%%%%%%%%%%%%%%%%%%%%%%%%%%%

\section{INTRODUCTION} \label{Sec:Intro}
% The concept of affordance holds significant importance in various robotic applications~\cite{gibson1966senses}, encompassing object recognition~\cite{thermos2017deep,hou2021affordance}, action anticipation~\cite{jain2016structural,roy2021action}, agent's activity recognition~\cite{vu2014predicting,qi2017predicting}, and object functionality understanding~\cite{jiang2022a4t}. These applications aim to utilize affordances to depict potential interactions between robots and their surrounding environment. For instance, in a general cutting task, one would seek to predict surfaces where the cutting is possible. These predictions enable robots to intelligently interact with knife-like objects, guiding them to use the knife's blade for specific purposes such as mincing meat or carving wood. Moreover, such affordance understanding facilitates interactive queries and visualizations, enhancing a person's comprehension. Nevertheless, this task is far from trivial, as robots must comprehend arbitrary correlations between objects, actions, and effects in complex and dynamic environments in real-time~\cite{min2016affordance}.

Intelligent robotic systems capable of interacting with objects and comprehending their affordances are of paramount importance across a wide array of real-world applications~\cite{liu2023survey}. These robotic applications encompass a diverse range of tasks, including object recognition~\cite{thermos2017deep,hou2021affordance}, action anticipation~\cite{jain2016structural,roy2021action}, agent's activity recognition~\cite{vu2014predicting,qi2017predicting}, and object grasping understanding~\cite{vuong2023graspanything}. In these tasks, the concept of affordance plays an important role as it refers to the potential actions or functionalities that an object can offer to its users. While affordance detection has received significant research interest in robotics, detecting object affordances poses significant challenges due to the inherent complexity and diverse shapes and functionalities of objects~\cite{min2016affordance}.
% \begin{figure}[t]
% 	\centering
% 	\includegraphics[width=.7\linewidth]{figures/intro/introduction.v23.pdf}
%  \vspace{1pt}
% 	\caption{\textbf{NEED TO RE-DRAW THIS FIGURE. THE CONCEPT IS NOT GOOD.}
%  We aim to improve the generalization capability of OpenAD~\cite{ngyen2023open}, which is limited to a predefined set of affordance labels. To achieve this, we are targeting 3D open-vocabulary Affordance Detection and leveraging 3D point cloud models trained on diverse datasets for understanding semantic parts of 3D objects. Our goal is to transfer this knowledge to the 3D affordance detection of target objects, particularly concerning out-of-domain 3D objects and novel affordances.}
% 	\label{fig:intro}
% \end{figure}

\begin{figure}
\centering
\subfigure[]{\label{fig:sub_intro1}\includegraphics[height=30mm]{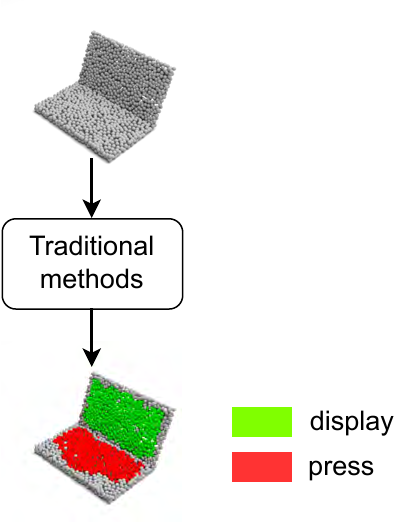}}
\subfigure[]{\label{fig:sub_intro2}\includegraphics[height=30mm]{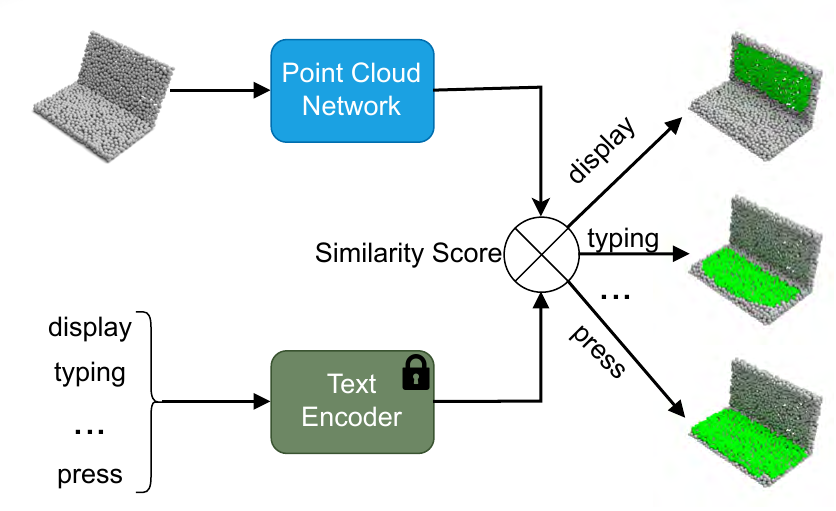}}
\subfigure[]{\label{fig:sub_intro3}\includegraphics[height=48mm]{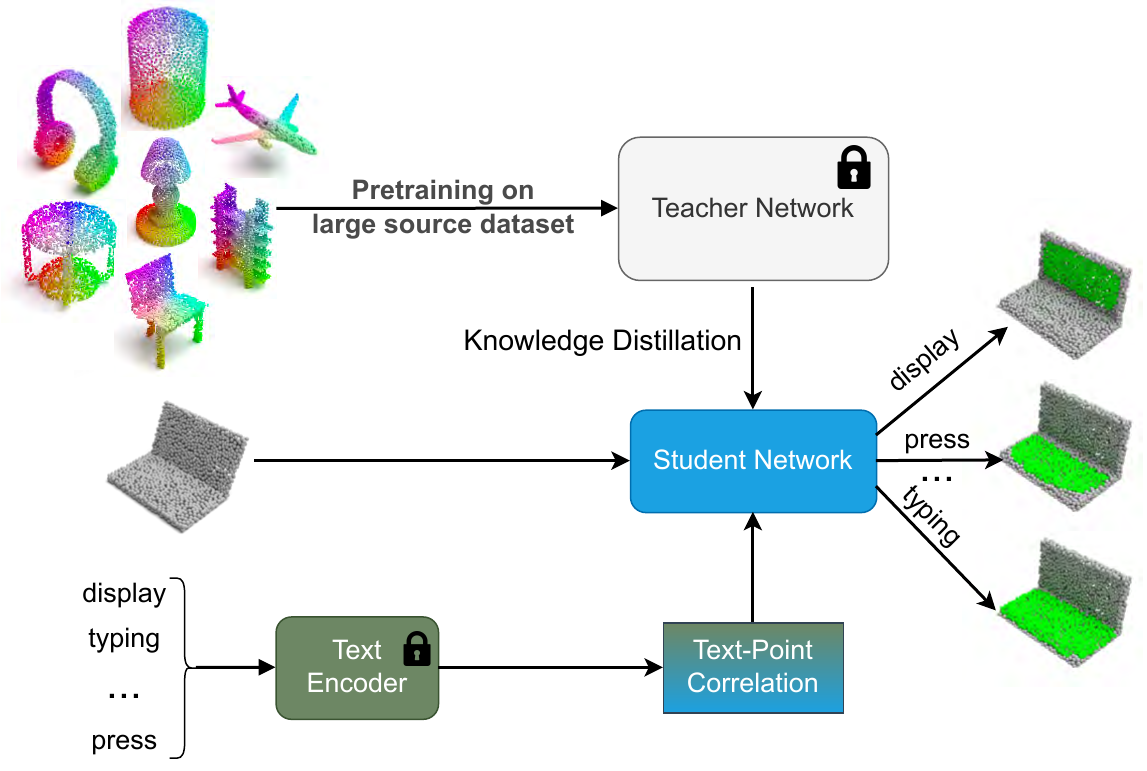}}
\vspace{1.5ex}
\caption{The comparison between: (a) traditional affordance detection methods, (b) OpenAD~\cite{ngyen2023open}, and (c) our proposed method. We leverage a point cloud teacher model and learn the text-point correlation to improve the open-vocabulary affordance detection results.}
\label{fig:intro}
\end{figure}

Classical affordance detection techniques have predominantly relied on traditional machine learning methods applied to images~\cite{hermans2011affordance}, texture-based cues~\cite{song2011visual}, relational affordance models~\cite{moldovan2014occluded} and human-object interactions~\cite{hassan2016attribute}. Deep learning, particularly Convolutional Neural Networks (CNNs)\cite{krizhevsky2017imagenet}, has also been employed for affordance-related tasks~\cite{mottaghi2017see,chuang2018learning,nguyen2016detecting,do2018affordancenet,luo2023leverage,nguyen2017object,li2019putting,pacheco2023one}. However, these methods confront challenges arising from the variability of visual information associated with object affordances despite their shared functionalities. Although leveraging 3D point clouds has gained popularity in robotics for supplying direct 3D object and environmental data, existing research~\cite{kokic2017affordance,deng20213d,iriondo2021affordance,mo2022o2o} has encountered limitations imposed by a \textit{fixed label set} tailored to specific tasks, limiting their support for broader or unsupervised inquiries. Furthermore, conventional approaches often encounter difficulties in capturing nuanced associations between localized point cloud regions and their corresponding labeled affordances~\cite{nguyen2017object}. 

%Additionally, an alternative approach~\cite{yang2023grounding}, grounded in linking 3D object affordances with 2D interactions in images, has made strides in 3D affordance detection. Nonetheless, this paradigm's reliance on the confines of 2D image space presents hurdles in capturing intricate 3D object-affordance relationships, resulting in the loss of spatial details, depth ambiguity, and viewpoint dependency that undermine accurate interpretation.

To overcome the fixed label set problem, the authors in~\cite{ngyen2023open} introduced an \textit{open-vocabulary} approach for 3D point cloud affordance detection that allows unrestricted natural language input, expanding the model's applicability. Despite promising strides, several limitations continue to challenge the effectiveness of existing methods for open-vocabulary affordance detection~\cite{ngyen2023open}. First, the inherent intricacies of 3D object shapes and the diverse range of potential affordances pose obstacles in the precise prediction and identification of object interactions~\cite{yu2021pixelnerf}. Second, the predicament of entirely novel or unseen affordances in real-world scenarios remains a formidable hurdle, necessitating the reinforcement of detection models' resilience and adaptability~\cite{deng20213d}. Finally, the intricate interplay between vision and language within the 3D environment mandates a more comprehensive comprehension of object-affordance relationships~\cite{peng2022openscene}. 
% To tackle these limitations, we propose the Interconnected Affordance Attention and Knowledge Transfer (IAAKT) framework. This comprises two essential sub-modules, the first being the Point-Point Cross Attention with Knowledge Distillation. Leveraging the effectiveness of knowledge distillation in diverse domains~\cite{distill_hinton,fitnets,kd_bert1,kd_bert2}, our method involves training a nimble student model to mimic the insights of a pre-trained, heavily parameterized teacher model. Attention-related knowledge is transferred from the teacher, enabling the fusion of local shape intricacies and the dynamics of interactions between global and local structures within 3D point clouds, ultimately enhancing affordance detection performance. Additionally, our study aims to thoroughly understand the intricate connection between 3D Open-Vocabulary Affordance and its interplay with vision and language. By introducing the Text-Point Cross Attention mechanism, we address the interpretability of 3D Open-Vocabulary Affordance. Utilizing the conventional attention mechanism~\cite{wang2018non}, we focus on point cloud regions in the context of affordances, bridging point regions and labeled affordances to identify latent associations. This strategy fosters an interpretable learning process that captures the nuanced interplay between vision and language, improving the point-text matching process and elucidating the relationships between objects and their corresponding affordances.

To tackle these limitations, we introduce a new open-vocabulary affordance detection in 3D point clouds using knowledge distillation and text-point correlation. Our method utilizes knowledge distillation to leverage 3D models pre-trained on large-scale datasets for the affordance detection task. Knowledge distillation fuses intricate local shape intricacies and dynamic interactions in 3D point clouds, reinforcing feature extraction without class-specific guidance. This enhancement, driven by attention knowledge transfer, enriches semantic comprehension in open-vocabulary affordance detection. Moreover, we introduce the text-point correlation to refine semantic connections between point cloud features and affordance labels. This approach, employing an established attention mechanism~\cite{wang2018non}, centers on relevant point cloud regions to strengthen the text-point relationships. The intensive experiment shows that our method demonstrates substantial improvements, achieving faster running time while improving $7.96$ mIOU score over the baselines. Ablation studies and qualitative results further validate the effectiveness of our approach and provide insights for future research directions.

Our main contributions are summarized as follows:
\begin{itemize}
    \item We propose a new approach to address the challenges of open-vocabulary affordance detection in 3D point clouds using knowledge distillation and text-point correlation.
    \item We intensively evaluate our method against prior methods and show its effectiveness in real-world robotic applications. Our code will be made available.
\end{itemize}

\section{Related Work} \label{Sec:rw}
\textbf{Affordance Detection.} Affordance detection is commonly approached as a pixel-wise labeling task, and numerous studies have focused on this area, as evident in~\cite{nguyen2016detecting,roy2016multi,nguyen2017object,do2018affordancenet,thermos2020deep,toan3DAP,chen2022cerberus,luo2022learning}.
The authors in~\cite{nguyen2017object}  developed a method to detect object affordances in real-world scenes by utilizing an object detector and dense conditional random fields.
In~\cite{do2018affordancenet}, the authors introduced a two-branch framework that simultaneously identifies multiple objects and their corresponding affordances from RGB images. Chen~\etal~\cite{chen2022cerberus} presented a multi-task dense affordance architecture.
More recently, Luo~\etal~\cite{luo2022learning} proposed a cross-view knowledge transfer framework to extract invariant affordances from exocentric observations.
%Other works have taken diverse approaches for their affordance detection tasks in images~\cite{hassan2016attribute,chen2015deepdriving}.
Hassan~\etal~\cite{hassan2016attribute} predicted high-level affordances by exploring the mutual contexts of humans, objects, and the surrounding environment, while Chen~\etal~\cite{chen2015deepdriving} learned meaningful affordance indicators for predicting actions in autonomous driving scenarios.

Affordance detection in the context of 3D point cloud data has been also the subject of extensive research~\cite{kokic2017affordance,deng20213d,iriondo2021affordance,mo2022o2o}. Kim~\etal~\cite{kim2014semantic} proposed a method that extracts geometric features from point cloud segments and employs logistic regression for affordance classification. Later, Kim and Sukhame~\cite{kim2015interactive} introduced a technique that voxelizes point cloud objects and generates an affordance map using interactive manipulation. Similarly, Kokic~\etal~\cite{kokic2017affordance} developed a system to model relationships between tasks, objects, and grasping. Iriondo~\etal~\cite{iriondo2021affordance} focused on detecting affordances for industrial bin-picking applications. Additionally, Mo~\etal~\cite{mo2022o2o} learned affordance heatmaps from object-object interactions. Yang~\etal~\cite{yang2023grounding} explores the task of linking 3D object affordances to 2D interactions in images. While these studies have made substantial contributions to affordance detection, the task of open-vocabulary affordance detection remains unexplored by these methods~\cite{ngyen2023open}.

\textbf{Open-Vocabulary Affordance Detection.} 
Recently, expansive vision language models have exhibited promising outcomes in robotic tasks~\cite{li2021language,xu2022simple}. Peng~\etal~\cite{peng2022openscene} harnessed a pre-trained textual encoder from CLIP~\cite{radford2021learning} to fuse 2D and 3D characteristics, aligning them with text embedding to address the challenge of open-vocabulary 3D scene comprehension. While these contributions have indeed propelled the advancements in affordance detection, they have yet to explicitly tackle the intricacies of open-vocabulary affordance detection. The authors in~\cite{ngyen2023open} introduced OpenAD, an open-vocabulary affordance detection method to identify a wide array of affordances in 3D point clouds. OpenAD effectively learns both textual and point-based affordance features, capitalizing on the semantic relationships among different affordances. However, a limitation of OpenAD lies in its generalization capability, particularly concerning out-of-domain 3D objects and novel affordances.

\begin{figure*}[ht]
	\centering
	\includegraphics[width=.9\linewidth]{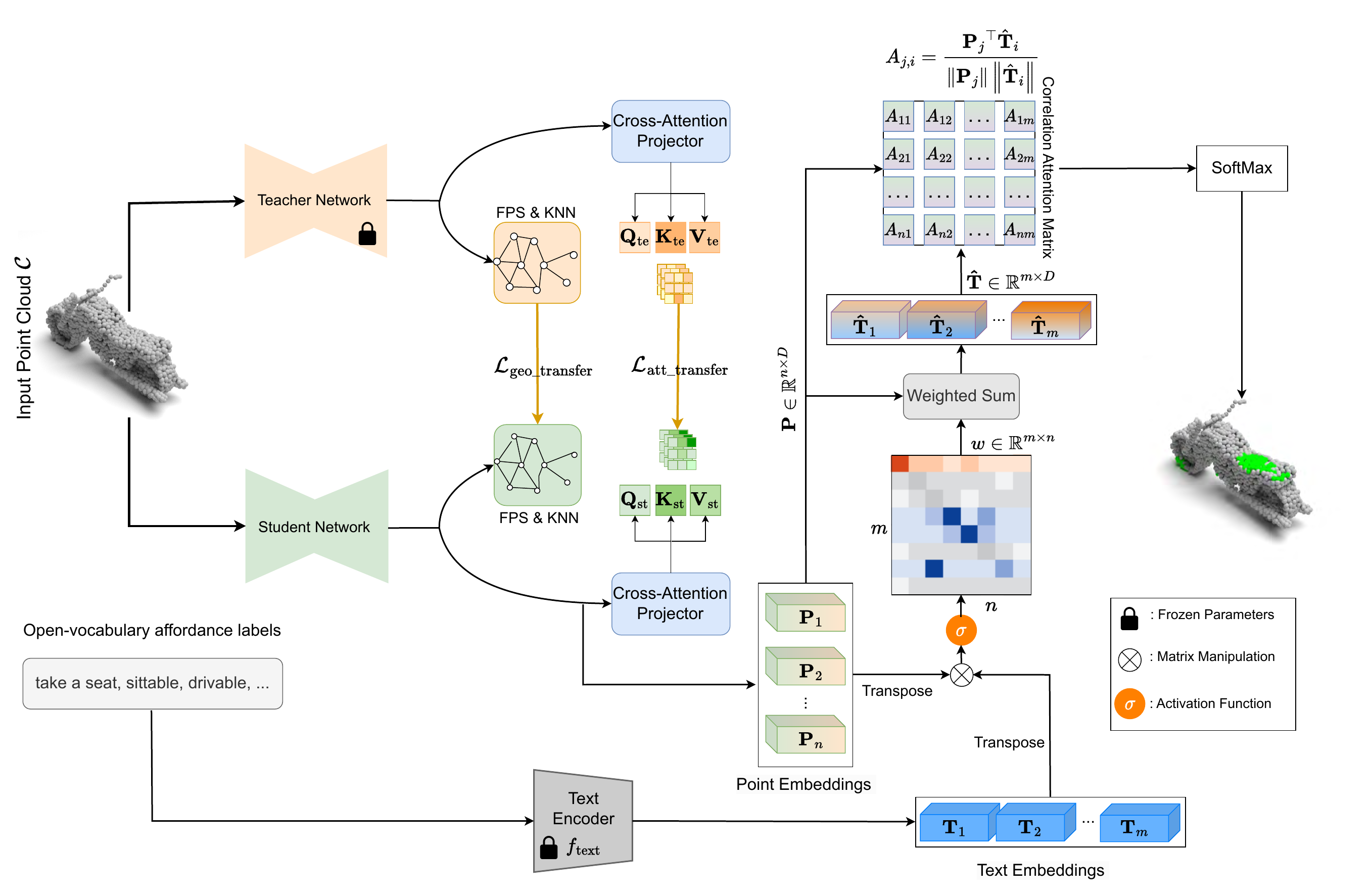}
 \vspace{0ex}
	\caption{An overview of our proposed open-vocabulary affordance detection method using knowledge distillation
and text-point correlation.}
	\label{fig:architecture}
\end{figure*}

\textbf{Knowledge Distillation.} Knowledge distillation entails the transfer of information from one network to another~\cite{hinton2015distilling}. Recently, there has been a shift towards cross-domain knowledge distillation~\cite{yao20223d,li2022cross,geng2023partmanip,zhang2022multi}, wherein knowledge is conveyed from a data-rich domain to one with limited diversity. For instance, Li~\etal~\cite{li2022cross} effectively employed cross-domain and cross-modal knowledge distillation to enhance 3D point cloud semantic segmentation across diverse scenarios. While cross-domain knowledge distillation has been extensively explored, its application to open-vocabulary affordance detection and its associated techniques remain relatively uncharted. Our study focuses on 3D open-vocabulary affordance detection, with the aim of harnessing diverse 3D point cloud models to transfer this knowledge to the open-vocabulary affordance detection task.

Differing from existing approaches that primarily focus on multi-modal student-teacher frameworks~\cite{li2022cross,peng2022openscene,huang2023joint}, our contribution centers on refining knowledge distillation for open-vocabulary affordance detection in 3D point clouds. Our method involves training a lightweight student model using insights from a well-parameterized teacher model. By transferring attention knowledge from teacher to student, our method enhances feature extraction at the point level, amplifying differentiation capabilities independently of class-specific guidance. %This refinement fosters improved semantic comprehension in open-vocabulary affordance detection models.
Additionally, we heighten semantic connections between point cloud affordances and labels through the established attention mechanism~\cite{wang2018non}. This strategy focuses on relevant point cloud regions to establish connections between point regions and labeled affordances, hence improving the point-text matching process and improving the final affordance detection results.

\section{Open-Vocabulary Affordance Detection} \label{Sec:method}

\subsection{Overview}
% We consider the task of open-vocabulary affordance detection in 3D point clouds by training a combined vision-language model. 
% Specially, we take into account an input point cloud $\mathcal{C} = \left \{\mathbf{p}_1, \mathbf{p}_2,...,\mathbf{p}_n  \right \}$ of $n$ unordered points, $\mathbf{p}_i \in \mathbb{R}^3$, $i=1,...,n$. 
% Each point is represented by its coordinate in Euclidean space. 
% The affordance labels are presented in a natural language form, i.e., $\mathcal{L} = \left \{\mathbf{l}_1, \mathbf{l}_2,..., \mathbf{l}_m \right \}$. Note that, in our open-vocabulary label setting, the number of labels $m$ can ideally be unlimited. The testing label set can differ from the training label set and can contain unseen affordance labels.

Following~\cite{ngyen2023open}, we address open-vocabulary affordance detection where the input cloud $\mathcal{C}$ contains $n$ unordered points $\mathbf{p}_i \in \mathbb{R}^3$, and the corresponding affordance labels are represented by natural language descriptions. The number of possible labels, denoted by $m$, can be unlimited, allowing adaptation to various affordance labels, even unseen ones during testing. Fig.~\ref{fig:architecture} shows an overview of our approach which has two branches: point-point attention with knowledge distillation and text-point correlation learning. % By employing knowledge distillation, we transfer knowledge from a larger, pre-trained teacher model to a smaller, more efficient student model, thereby enhancing affordance detection. Moreover, our introduced text-point correlation effectively establishes correlations between visual features of the point cloud and text embedding features, improving the understanding of 3D object-affordance relationship.

\subsection{Point-Point Attention with Knowledge Distillation}
We utilize a teacher model (pre-trained on large datasets~\cite{chang2015shapenet}) to transfer its knowledge to the student model via a cross-attention distillation
mechanism that minimizes the dissimilarity between the student and teacher attention maps. In particular, the point cloud network processes $n$ input points, resulting in an embedding vector for each point, represented as $\mathbf{P}_1, \mathbf{P}_2,...,\mathbf{P}_n\in\mathbb{R}^D$, for both the student and teacher models. Following~\cite{yang2023geometry}, to model the geometry of the point cloud, we first use Farthest Point Sampling (FPS) on the input point cloud to uniformly sample r-proportional points (the number is $\left\lfloor Z=r*n \right\rfloor$) as anchors $\left\{ \mathcal{C}^{a} \right\}_{a=1}^{Z}$. We then calculate the Euclidean distance between each point and the anchors and apply K-Nearest Neighbors (KNN) to sample the nearest $K$ points $\mathcal{C}^{a,k}, k \in \mathrm{\mathcal{N}\text{(a)}}$ to form local areas reflecting the geometric structures. Based on this, we represent the point-wise relative relationships $\mathcal{R}^a$ within the geometric neighbors, which contain the structured knowledge for migration and can be formulated as:
% Specifically, the $n$ input points are plugged into the point cloud network $f_{\rm pc}\left ( \cdot\right )$ producing an embedding vector for every input point. The point cloud network $f_{\rm pc}\left ( \cdot\right )$ produces a set of $n$ vectors $\mathbf{P}_1, \mathbf{P}_2,...,\mathbf{P}_n\in\mathbb{R}^D$, similarity for student model and teacher model.
% To model the geometry of the point cloud, we first apply the Farthest Point Sampling (FPS)~\cite{moenning2003fast} on the input point cloud to uniformly sample r-proportional points (the number is $\left\lfloor Z=r*s \right\rfloor$) as anchors $\mathrm{\left\{\mathrm{ \mathbf{P}}_{a}^{n} \right\}}_{a=1}^{Z}$. We calculate the 2 distance between each point and anchors using $xyz$ coordinates, and apply KNN to sample the nearest $K$ points $\mathrm{\mathbf{P}}_{n}^{a,k}, k \in \eta(a)$ to form areas reflecting the local geometric structures. On this basis, we represent the point-wise relative relationships $R^a$ within the geometric neighbors, which carry the crucial structured knowledge for migration and can be formulated as:
\begin{equation}
    \mathcal{R}^a = \frac{1}{K}\sum_{k \in \mathcal{N}\text{(a)}}\left( \mathrm{\mathbf{p}}_{n}^{a,k} - \mathrm{\mathbf{p}}_{n}^{a} \right)\oplus \left( \mathrm{\mathbf{P}}_{n}^{a,k} - \mathrm{\mathbf{P}}_n^{a} \right),
\label{eq: kl_local_local}    
\end{equation}
where $\mathbf{p}_n$ are $xyz$ coordinates of points in the set $\mathcal{C}$ of $n$ input points, $\mathbf{P}_n$ are the embedding vector for every input point $\mathbf{p}_n$, $\oplus$ indicates the concatenation operation. The point-wise feature relations of the teacher and student model can be expressed as $\mathcal{R}_{\rm te}^{a}$ and $\mathcal{R}_{\rm st}^{a}$ respectively. We transfer the knowledge of the teacher to the student via the MSE loss:
\begin{equation}
   \mathcal{L}_\text{geo\_transfer} = \frac{1}{Z}\sum_{a=1}^{Z}\left\| \mathcal{R}_{\rm te}^{a} - \mathcal{R}_{\rm st}^{a} \right\|,
\label{eq: kl_local_local_mseloss}    
\end{equation}

Subsequently, the Cross-Attention Projector transforms the feature space of both the student and teacher point clouds into the transformer attention space. It is achieved by mapping point features into query, key, and value matrices%, thereby replicating the attention mechanism
. The self-attention~\cite{vaswani2017attention} is used to capture local relationships among objects by first generating query, key, and value embeddings from the feature matrix $\mathbf{P} \in \mathbb{R}^{n \times D}$, where $\mathbf{Q}=\mathbf{P}W_Q$, $\mathbf{K}=\mathbf{P}W_K$, and $\mathbf{V}=\mathbf{P}W_V$. Here, $W_Q$, $W_K$, and $W_V \in \mathbb{R}^{d \times d_h}$ are trainable parameters, with $d$ denotes the query size and $d_h$ is the output embedding dimension. We compute matrix representations $\mathbf{Q}_{\rm st}, \mathbf{K}_{\rm st}, \mathbf{V}_{\rm st}$ to model the attention in the student's space, and matrix representations $\mathbf{Q}_{\rm {te}}, \mathbf{K}_{\rm te}, \mathbf{V}_{\rm te}$ for the teacher's attention space.
%The matrices $\mathbf{Q}_{S}, \mathbf{K}_{S}, \mathbf{V}_{S}$ are computed and aligned to mirror the query $\mathbf{Q}_{T}$, key $\mathbf{K}_{T}$, and value $\mathbf{V}_{T}$ of the teacher's attention space. Consequently, within the attention space, t
The self-attention mechanism for the student is computed as:
\begin{flalign}
	\begin{aligned}
		\label{eqn:self_attn}
		\Omega_{\rm st} = \text{softmax}(\dfrac{\mathbf{Q}_{\rm st}(\mathbf{K}_{\rm st})^\mathrm{\top}}{\sqrt{d}})\mathbf{V}_{\rm st},
	\end{aligned}
\end{flalign}
The calculation of $\Omega_{\rm te}$ for the teacher is similar. Hence, we can minimize the distance between the attention maps of the teacher and the student to guide the student network using the following objective function:
\begin{equation}
\label{eqn:attn_distill_loss}
	\mathcal{L}_{\rm att\_transfer} = ~\text{MSE}(\Omega_{\rm te} ,\Omega_{\rm st}),
\end{equation}
where $\text{MSE}(\cdot)$ is the mean square error function.

\subsection{Text-Point Correlation}
% \begin{figure}[ht]
% 	\centering
% 	\includegraphics[width=0.9\linewidth]{figures/method/attention_point_Text_v1.pdf}
%  \vspace{1ex}
% 	\caption{Text-Point Correlation Cross Attention:. 
%  }
% 	\label{fig:attention_textPoint}
% \end{figure}
% In our approach, we represent point and text-affordance features as matrices: $\mathbf{P}=[\mathbf{P}_1, \mathbf{P}_2,...,\mathbf{P}_n]\in\mathbb{R}^{n \times D}$ and $\mathbf{T} = [\mathbf{T}_1, \mathbf{T}_2,...,\mathbf{T}_m] \in\mathbb{R}^{m\times D}$, respectively. To achieve a better understanding of the interaction between vision and language, it is essential to capture the underlying attention connection between points and text-affordances. We specifically highlight attention points within the point cloud features space concerning each text-affordance. To be specific, we first preassign attention by computing the similarity score between each point and text-affordance using cosine similarity, and normalize similarity scores of each text-affordance with respect to query point into [0,1] using softmax, attention on \begin{math}j\end{math}-th text-affordance is denoted as \begin{math} w_{ij}\end{math}: 
In our approach, following previous works~\cite{ngyen2023open}, we use the text encoder from CLIP~\cite{radford2021learning}, which produces $m$ word embeddings $\mathbf{T} = [\mathbf{T}_1, \mathbf{T}_2,...,\mathbf{T}_m] \in\mathbb{R}^{m\times D}$ for text-affordance labels. To understand the interaction between vision and language, we focus on the correlation in the feature space of each text-affordance. Let $\mathbf{T}_{i}$ and $\mathbf{P}_{j}$ refer to the feature representation of the \begin{math}i\end{math}-th affordance query and \begin{math}j\end{math}-th points, respectively. We compute the correlation by calculating the $\mathbf{P}_j$’s attention weight with respect to text-affordance $\mathbf{T}_i$ as:
% first measuring similarity scores between each point and text-affordance using cosine similarity, then normalizing them using softmax function. This attention for the \begin{math}j\end{math}-th text-affordance is denoted as \begin{math} \mathbf{w}_{j}\end{math}:
\begin{equation}
    w_{i,j}=\sigma (\mathbf{T}_{i}^{\top}\mathbf{P}_{j}), i\in [1,m], j\in [1,n],
\end{equation}
where $\sigma$ is the activation function.
%, $w_{i,j}$ is $\mathbf{P}_{j}$’sattention weight with respect to $\mathbf{T}_{i}$.
% \begin{math}\sigma\end{math} denotes softmax activation. We calculate the similarity matrix $H$ between point and text-affordance as:
% \begin{equation}
%     H_j = \text{tanh}(\mathbf{P}_{i}^{\top}K\mathbf{T}_{j}), i\in [1,n], j\in [1,m],
% \end{equation}
% where $K \in\mathbb{R}^{D \times D}$ is weight matrix.
% During this process, relevant text-affordance can be paid more attention, but irrelevant text-affordance  also contribute to shared semantics between points and target affordance. The attention for relevant text-affordance will be reassigned:
% \begin{equation}
% \mathbf{w}_{j}^{'}=\frac{\mathbf{w}_{j}H_{j}}{\sum_{j=1}^{m} \mathbf{w}_{j}H_{j}}, j\in [1,m].
% \end{equation}

The attention features $\mathbf{\hat{T}}_i$ for the affordance text $\mathbf{T}_i$, is defined from the point feature $\mathbf{P}_{j}$ and a weighted summation of keypoint features, as in the following equation:
% The text-affordance $\mathbf{T}_i$’s attended information derived from the point feature  $\mathbf{P}_{j}$ is defined as the attention feature $\mathbf{\hat{T}_i}$ using the weighted sum of key points in the following equation:
% \begin{equation}
%     \mathbf{\hat{T}}_i=\sum_{j \in n}(w_{i,j}\mathbf{P}_{j}).
% \end{equation}
% \begin{equation}
%     \hat{\mathbf{T}_i}= \sigma (w_{i,j}\times\mathbf{P}_j), i\in [1,m], j\in [1,n],
% \end{equation}
% \begin{equation}
%     \hat{\mathbf{T}_i}= \dfrac{\sigma (w_{i,j}\times\mathbf{P}_j)}{\sum_{j=1}^{n}\sigma(w_{i,j}\mathbf{P}_j)}, i\in [1,m], j\in [1,n],
% \end{equation}

\begin{equation}
    {\mathbf{\hat{T}}_i}= \dfrac{\sum_{j=1}^{n}\sigma(w_{i,j}\mathbf{P}_j)}{\sum_{j=1}^n{w_{i,j}}}, i\in [1,m], j\in [1,n],
\end{equation}
%where $\sigma$ is activation function.

% The reassigned attention will be redirected towards all pertinent text-affordances by performing an element-wise product with their representations in a \begin{math}D\end{math}-dimensional space. 

% The shared semantic information with the text-affordance is derived from the point features, computed as a weighted combination of the relevant text-affordance, denoted as \begin{math}\mathbf{\hat{T}_j}=\mathbf{w}_{j}^{'}\mathbf{T}_{j}, j\in [1,m]\end{math}. The weighting in this combination is determined by the learned attention.
The overall relevance score for the text-point correlation attention matrix is then computed as follows:
\begin{equation}
    A_{j,i} =\frac{\mathbf{P}_{j}^{\top}\mathbf{\hat{T}}_{i}}{\left \| \mathbf{P}_{j} \right \|\left \| \mathbf{\hat{T}}_{i} \right \|},j\in [1,n], i\in [1,m].
\end{equation}
Following~\cite{ngyen2023open}, the point-wise softmax output of a single point $i$ is then computed in the form:
\begin{equation}
    {s}_{j,i}= 
    \dfrac{\exp\left ({A}_{j,i}/\tau\right )}
    {\sum_{k=1}^{m}\exp({{A}_{k,i}}/\tau)}\:\:,
\label{eq: S}    
\end{equation}
where $\tau$ is a learnable parameter~\cite{wu2018unsupervised}. We aim to maximize the value of the entry ${s}_{j,i}$ that is the attention correlation of $\mathbf{P}_j$ and the text attention embedding $\mathbf{\hat{T}}_i$ corresponding to the ground-truth label $i = y_i$.
This can be accomplished by optimizing the weighted negative log-likelihood loss of the point-wise \text{softmax} output over the entire point cloud in the form:
\begin{equation}
    { \mathcal{L_\text{point\_wise}}= -\sum_{j=1}^n \mathcal{\omega}_{y_i}\log {s}_{j,{y_i}}} \:\:,
\end{equation}
where $\mathcal{\omega}_{y_i}$ is the weighting parameter to the imbalance problem of the label classes during the training.

Finally, the overall training objective is the combination of both loss terms $\mathcal{L}_{\rm total}$:
\begin{align}
\mathcal{L}_{\rm total}  &=  \mathcal{L}_{\text{point\_wise}}+\lambda_{\text{a}} \mathcal{L}_{\rm att\_transfer} + \lambda_{\text{t}} \mathcal{L}_{\rm geo\_transfer} 
\end{align}
where $\lambda_\text{a}, \lambda_\text{t}$  is hyper-parameter to balance loss.

\section{Experiments} \label{Sec:exp}
%In this section, we perform several experiments to validate the effectiveness of our method. We start with a zero-shot detection setting to verify the ability of our method to generalize to previously unseen affordances. Secondly, we present our's notable qualitative results together with visualizations. Finally, we conduct additional ablation studies to further investigate other aspects of 3D Open-Vocabulary Affordance Detection.

\subsection{Experimental Setup}
\textbf{Dataset.}
We employ the 3D AffordanceNet dataset~\cite{deng20213d} and its open affordance labels by~\cite{ngyen2023open} for our experiments. This dataset is the large-scale dataset for affordance detection using 3D point clouds, containing $22,949$ instances across $23$ object categories. %, and $37$ open-vocabulary affordance labels. %. As in other zero-shot setups~\cite{cheraghian2019zero,cheraghian2020transductive,michele2021generative}, to ensure robustness, we expand the label classes by re-labeling the 3D AffordanceNet dataset with an additional $18$ affordance classes, including the background as a class. This results in a total of $37$ affordance labels. 
Following~\cite{deng20213d,ngyen2023open}, we evaluate our approach on two tasks: full-shape and partial-view. The partial-view setup is particularly relevant in robotics, as it reflects the limited observation capability of robots, where only a partial view of the object's point cloud is available.

\textbf{Baselines and Evaluation Metrics.} 
We conduct a comparative analysis of our method with recent approaches in zero-shot learning for affordance detection in 3D point clouds, including ZSLPC~\cite{cheraghian2019zero}, TZSLPC~\cite{cheraghian2020transductive}, 3DGenZ~\cite{michele2021generative}, and OpenAD~\cite{ngyen2023open}. %To ensure a fair comparison, we replace the original text encoders in these baselines, which employed GloVe~\cite{pennington2014glove} or Word2Vec~\cite{mikolov2013distributed}, with CLIP, a more powerful model. Additionally, for ZSLPC~\cite{cheraghian2019zero} and TZSLPC~\cite{cheraghian2020transductive}, we modify their classification heads to suit the segmentation task. 
To evaluate the results, we utilize three metrics commonly used in related studies~\cite{ngyen2023open}, namely, mIoU (mean IoU over all classes), Acc (overall accuracy over all points), and mAcc (mean accuracy over all classes).  During training, we keep the text encoder and pre-trained teacher model fixed, then train the rest end-to-end. Point cloud size is fixed at $n = 2048$ and $D$ at $512$ as in~\cite{ngyen2023open}. The hyperparameters $\tau$, $\lambda_\text{a}$ and $\lambda_\text{t}$ are set to $\ln{(1/0.07)}$, $0.9$ and $0.7$, respectively. Finally, we use Adam optimizer with $\alpha = 10^{-3}$ and $\gamma = 10^{-4}$ for $200$ epochs on an NVIDIA A100 40GB and batch size of $16$. %, $\tau$ is initialized as $\ln{(1/0.07)}$. 

\begin{table}[ht]
\caption{Zero-shot Open-vocabulary detection results}
\label{tab: zero-shot task result}
\vskip 0.15in
\begin{center}
\resizebox{\columnwidth}{!}{%
\begin{tabular}{llcccccc}
\toprule
Task & Method  &  mIoU & Acc & mAcc&Params& CPU(s) & GPU(s) \\
\midrule
\multirow{5}{*}{\rotatebox{90}{Full-shape}} & TZSLPC\cite{cheraghian2020transductive} & 3.86 & 42.97 & 10.37 & 1.7M&0.75& 0.13 \\
& 3DGenZ~\cite{michele2021generative} & 6.46 & 45.47 & 18.33 & 1.79M &0.76 &0.14 \\
& ZSLPC~\cite{cheraghian2019zero} & 9.97 & 40.13 & 18.70 & 1.96M&0.82 &0.16  \\
% & OpenAD (DGCNN) (ours) & 10.88 & 45.21 & 15.40 \\
& OpenAD~\cite{ngyen2023open} & 14.37 & 46.31 & 19.51  & 1.8M & 0.77 & 0.14 \\
& \bf{Ours} & \bf{22.33} & \bf{49.72} & \bf{34.29} &\bf{0.58M} & \bf{0.43} & \bf{0.12} \\
\midrule
\multirow{5}{*}{\rotatebox{90}{Partial-view}} & TZSLPC~\cite{cheraghian2020transductive} & 4.14 & 42.76 & 8.49 & 1.7M&0.75 &0.13 \\
& 3DGenZ~\cite{michele2021generative} & 6.03 & 45.24 & 15.86& 1.79M &0.76&0.14 \\
& ZSLPC~\cite{cheraghian2019zero} & 9.52 & 40.91 & 17.16& 1.96M&0.82&0.16 \\
% & OpenAD (DGCNN) (ours) & 11.19 & 44.21 & 16.43 \\
& OpenAD~\cite{ngyen2023open}  & 12.50 & 45.25 & 17.37 & 1.8M & 0.77 &0.14 \\
& \bf{Ours}  &\textbf{20.48} & \textbf{48.72} & \textbf{32.86} &\bf{0.58M}  &\bf{0.43} &\bf{0.12}\\
\bottomrule
\end{tabular}
}
\end{center}
\vskip -0.1in
\end{table}

\subsection{Quantitative Results}
% \lipsum[1]
% We present several examples to demonstrate the generality and flexibility of OpenAD. Primarily, we use objects from the 3D AffordanceNet~\cite{deng20213d} for our visualizations. We also select objects from the ShapeNetCore dataset~\cite{chang2015shapenet} to analyze the capability of OpenAD to generalize to unseen object categories and new affordance labels.
The comparison results of evaluation metrics are shown in Table~\ref{tab: zero-shot task result}. As can be seen, our approach achieves superior results on both tasks and all three evaluation metrics.
Particularly on the full-shape task, our method outperforms the runner-up model (OpenAD) by a substantial margin of
7.96\% in mIoU. Additionally, our method shows significant superiority over the other approaches, surpassing OpenAD
by 14.78\% in mAcc and by 3.41\% in Acc.

In terms of operational efficiency, our method also significantly outperforms other baselines. On CPU, we achieve a 1.5 times speedup. Moreover, the number of parameters during inference is scaled down by 3 times. Importantly, these efficiency gains do not compromise our method's performance superiority compared with other methods.

\begin{figure}[t]
	\centering
	\includegraphics[width=1.\linewidth]{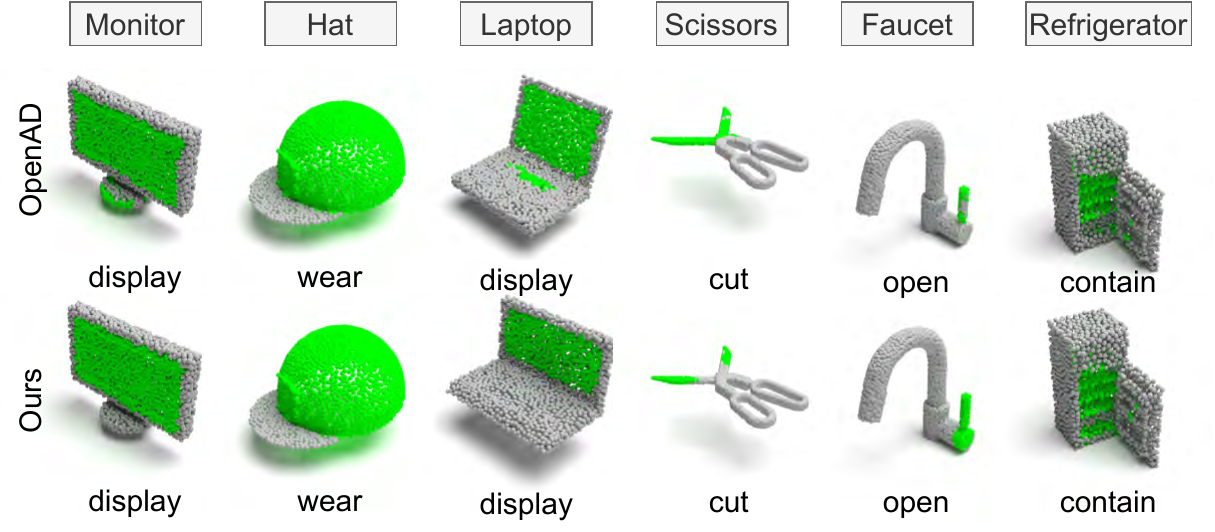}
 \vspace{1.2ex}
	\caption{The visualization of our method and OpenAD~\cite{ngyen2023open} when detecting seen affordances.}
	\label{fig:seen_affordances}
\end{figure}

\begin{figure}[t]
	\centering
	\includegraphics[width=1.\linewidth]{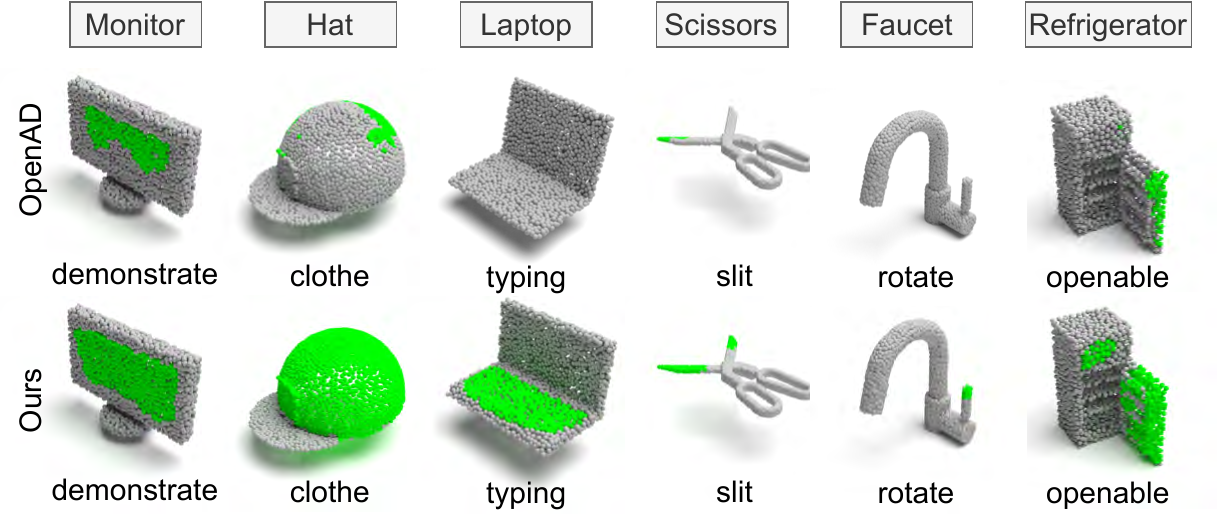}
 \vspace{1.2ex}
	\caption{The visualization of our method and OpenAD~\cite{ngyen2023open} when detecting unseen affordances that do not exist in the training set.}
	\label{fig:unseen_affordances}
\end{figure}

\subsection{Quantitative Results}
%\textcolor{blue}{done}\textbf{Visualization.} To illustrate the limitation of methods that lock geometrics with specific unseen affordance categories, we conduct an experiment to compare one of these methods~\cite{ngyen2023open} with ours. 

\begin{figure}[t]
	\centering
	\includegraphics[width=1.\linewidth]{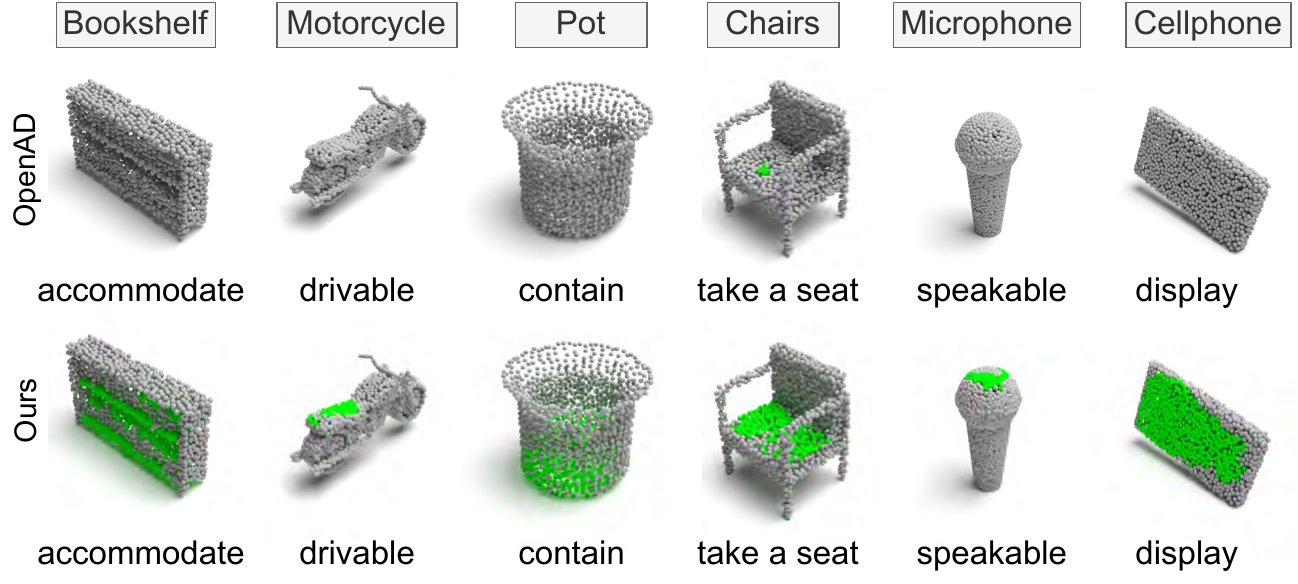}
 \vspace{1pt}
	\caption{Results on unseen object categories and unseen affordances.}
	\label{fig:unseen_objects}
\end{figure}

\begin{figure}[t]
	\centering
	\includegraphics[width=1.\linewidth]{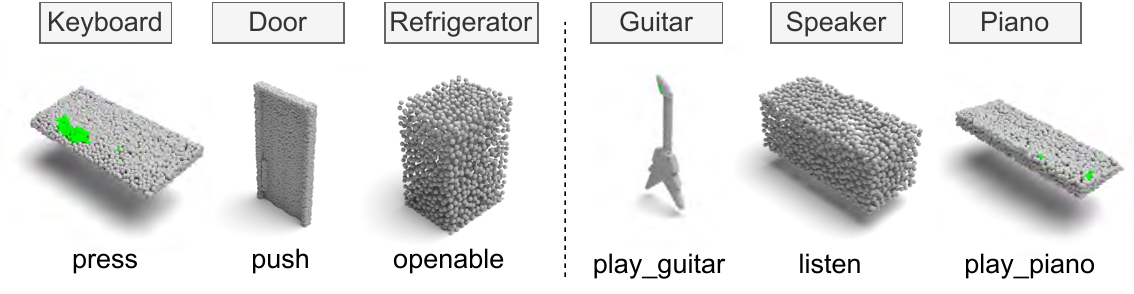}
 \vspace{-1pt}
	\caption{Failure cases of our method.}
	\label{fig:fail_case}
\end{figure}

\textbf{Seen Affordances.} Fig.~\ref{fig:seen_affordances} presents a visual comparison between our method and OpenAD~\cite{ngyen2023open}. While detecting seen affordances is relatively straightforward, this illustration underscores OpenAD's struggles in accurately identifying known affordance areas when they overlap with other regions. For instance, for objects like \texttt{monitor}, \texttt{laptop}, \texttt{faucet}, and \texttt{refrigerator}, OpenAD often misidentifies non-affordance areas as affordance zones. In contrast, our approach consistently delivers precise results for affordance regions, avoiding confusion with other areas. %This observation underscores the efficacy of our method and its potential applicability in scenarios requiring adaptable and accurate focus on affordance regions.

\textbf{Unseen Affordance.}  Fig. \ref{fig:unseen_affordances} shows the comparison with unseen affordance inputs. While it is more challenging to detect unseen affordances, this figure illustrates that our method still achieves better results compared to OpenAD~\cite{ngyen2023open}. For instance, when considering a \texttt{laptop} object, the baseline method struggles to distinguish between the keyboard and screen areas. In contrast, our method adeptly addresses these challenges, exhibiting an enhanced ability to discern subtle differences among affordance regions. %This improvement is attributed to the proficient utilization of the Text-Point Cross Attention mechanism, coupled with the rich knowledge transferred from the teacher model to the affordance model, thereby enriching the understanding of relationships between point clouds and affordances. 
%Our method excels in detecting both seen and unseen affordances, highlighting its adaptability and discrimination.

%We conduct a rigorous assessment of our method's robustness when confronted with unseen affordances that were not part of the training set. The results, as depicted in Fig. \ref{fig:unseen_affordances}, unequivocally demonstrate the exceptional resilience of our approach when compared to the base method~\cite{ngyen2023open}. Notably, our method exhibits remarkable generalization capabilities, even in scenarios where the training set includes labeled affordances. In such cases, our model detects affordance regions that exhibit a higher level of generality compared to the baseline method. For instance, in the case of the laptop object, the baseline method struggles to effectively differentiate between the keyboard area and the screen area, highlighting one of its inherent limitations. In contrast, our method successfully addresses these challenges and demonstrates an improved capacity to distinguish subtle variations among affordance regions. This advancement is accomplished by effectively leveraging the Text-Point Cross Attention mechanism, enabling a more profound comprehension of the intrinsic relationships between point cloud regions and their corresponding labeled affordances. As a result, our method not only surpasses the baseline in detecting unseen affordances but also exhibits superior generalization capabilities when encountering familiar objects with labeled affordances.

\textbf{Unseen Objects.} We assess the robustness of our method in dealing with new object categories, a key evaluation criterion. Our approach outperforms the baseline model, demonstrating superior adaptability to previously unseen objects, as shown in Fig. \ref{fig:unseen_objects}. This showcases our method's effectiveness and its potential value in scenarios requiring adaptability to novel objects.
%We delve into the robustness of our proposed method in the face of unseen object categories, a critical aspect in the evaluation of our approach. We particularly focus on assessing the performance of our method in comparison to the baseline model. Unseen object categories pose a significant challenge, as they necessitate the detection of affordances on objects that were not present during the training phase. This scenario reflects real-world applications where novel objects might need to be interacted with intelligently. Through rigorous experimentation, we showcase that our method exhibits superior robustness when confronted with such unseen object categories, as compared to the baseline~\cite{ngyen2023open}, as depicted in Fig. \ref{fig:unseen_objects}. This not only underscores the effectiveness of our approach but also highlights its potential applicability in scenarios requiring adaptability to previously unencountered objects.
% \lipsum[1]

\textbf{Failure Cases.} While our method significantly enhances generalizability to unseen affordances and objects, challenges persist with highly semantic affordances and unfamiliar objects featuring intricate geometric structures. In some cases, the semantic information varies when these objects are placed in diverse contexts, as illustrated in Fig.~\ref{fig:fail_case}. For instance, the resemblance between a \texttt{keyboard} and a \texttt{piano}, or a \texttt{refrigerator} and a \texttt{speaker}, presents difficulties due to their similar box-like shapes and planes. This makes it challenging to pinpoint the affordance zones.

\begin{figure}[t]
	\centering
	\includegraphics[width=1.\linewidth]{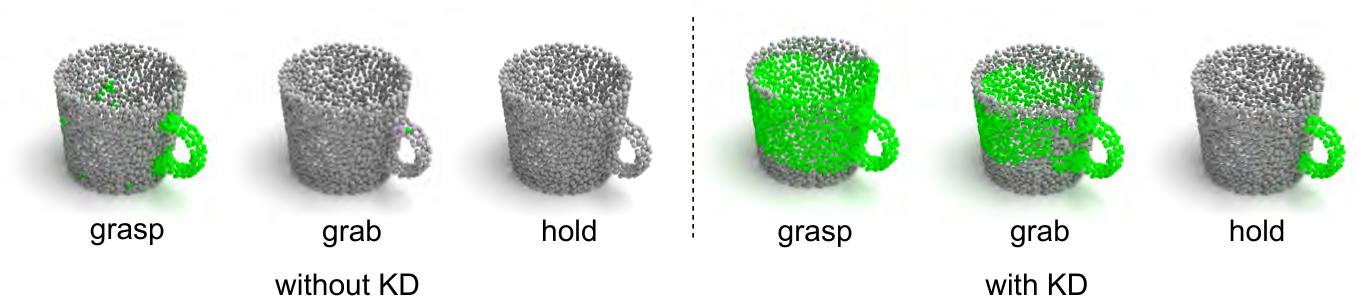}
 \vspace{1pt}
	\caption{Effectiveness of Knowledge Distillation (KD).}
	\label{fig:vls_score_unseen_objects}
\end{figure}

\begin{figure}[t]
	\centering
	\includegraphics[width=.95\linewidth]{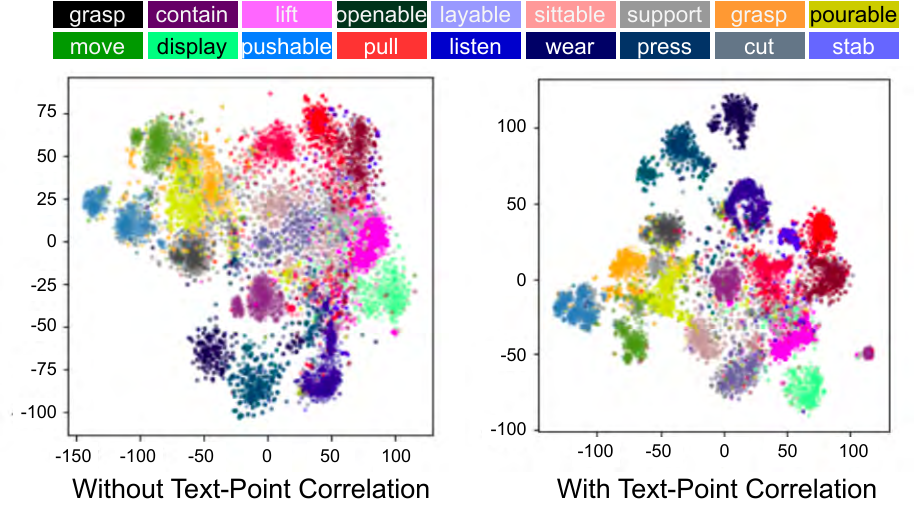}
 \vspace{1ex}
	\caption{Effectiveness of Text-Point Correlation though t-SNE visualization of the feature maps at the last stage of student point cloud network for each related affordance.}
	\label{fig:cross-similarity matrix_score}
\end{figure}

\begin{table}[!ht]
\caption{Effectiveness of Knowledge Distillation (KD) and Text-Point Correlation (TPC) in our method.}
\label{table:ablation}
\vskip 0.15in
\begin{center}
\begin{tabular}{lccccc}
\toprule
Task & KD &TPC  &  mIoU & Acc & mAcc \\
\midrule
Full-shape & & & 14.37 & 46.31 & 19.51 \\
& \checkmark& & 21.19 & 48.29 & 32.32 \\
& & \checkmark & 21.13 & 48.33 & 32.45 \\
% & OpenAD (DGCNN) (ours) & 10.88 & 45.21 & 15.40 \\
& \checkmark & \checkmark & \textbf{22.33} & \textbf{49.72} &\textbf{34.29} \\
\midrule
Partial-view & & & 12.50 & 45.25 & 17.37 \\
& \checkmark& & 19.06 & 47.65 & 28.13 \\
& &\checkmark & 19.72 & 48.02 & 29.11 \\
% & OpenAD (DGCNN) (ours) & 11.19 & 44.21 & 16.43 \\
&\checkmark&\checkmark  &\textbf{20.48} & \textbf{48.72} & \textbf{32.86} \\
\bottomrule
\end{tabular}
\end{center}
\vskip -0.1in
\end{table}

\subsection{Knowledge Distillation and Text-Point Attention Analysis}

% \textbf{Will language-driven architecture affect the detection results?} In this work, we mainly focus on jointly learning a vision-language model to improve the generalization in downstream robotic tasks, and do not aim for improving the accuracy of the traditional affordance detection tasks. This leads to the question that whether our language-driven model affects the accuracy of the traditional affordance detection task. To verify this, we train our OpenAD on the original 3D AffordanceNet with its original label set and training split, and compare our result with other state-of-the-art methods, including PointNet++~\cite{qi2017pointnet++}, Dynamic Graph CNN (DGCNN)~\cite{wang2019dynamic} and Point Transformer~\cite{zhao2021point}. For PointNet++ and DGCNN, we follow similar designs in~\cite{deng20213d} and change the final classifier to a linear layer detecting the affordance classes. For Point Transformer, we apply the same architecture in~\cite{zhao2021point}. Table~\ref{tab: original task result} presents the results of all methods on 3D AffordanceNet~\cite{deng20213d}. From this table, we find that OpenAD performs competitively when compared to other methods. Therefore, we can conclude that while our method is designed for a different purpose, it still can be used as a strong benchmark for closed-set affordance detection.

\textbf{Effectiveness of Knowledge Distillation.} The impact of knowledge distillation is demonstrated in Table~\ref{table:ablation}. Additionally, we visually assess the effectiveness of Knowledge Distillation in Fig.~\ref{fig:vls_score_unseen_objects}, where the left side illustrates the student embeddings without Knowledge Distillation (KD), and the right side shows the student representations learned with KD. These results illustrate that Knowledge Distillation directs the model's attention towards interactive regions, facilitating the extraction of interaction contexts.

\textbf{Effectiveness of Text-Point Correlation.}
Table~\ref{table:ablation} reports the impact of the Text-Point Correlation (TPC) in our method. Additionally, we demonstrate the representation of the input text and learned
embeddings in the latent space via t-SNE visualizations~\cite{van2008visualizing} in Fig.~\ref{fig:cross-similarity matrix_score}. The results show that without TPC, the decision boundaries for most of the affordance are obscure and difficult to distinguish during the training. On the other hand, applying TPC increases both the accuracy and learned features of the network.  %, which leads to harder decision making for the network.

\begin{table}[h]
\caption{Teacher models comparison}
\label{tab: choose teacher}
\vskip 0.15in
\begin{center}
\begin{tabular}{llccc}
\toprule
Task & Method  &  mIoU & Acc & mAcc \\
\midrule
Full-shape & {Point Transformer}~\cite{zhao2021point} & 40.32 & 64.38 & 65.22 \\
& {PointNet++}~\cite{qi2017pointnet++} & \bf{42.47} & \bf{68.60} & \bf{66.55} \\
& DGCNN~\cite{wang2019dynamic} & 41.83 & 67.43 & 64.41 \\
& PointMAE~\cite{pang2022masked} & 40.17 & 63.52 & 64.28 \\
&PAConv~\cite{xu2021paconv} & 38.52 & 58.14 & 59.48 \\
\midrule
Partial-view & {Point Transformer}~\cite{zhao2021point}  & 40.18 & 64.29 & 64.52 \\
& {PointNet++}~\cite{qi2017pointnet++} &\bf{41.94}  &\bf{68.72}  & \bf{66.58} \\
& DGCNN~\cite{wang2019dynamic}  & 41.52 & 67.01 & 63.22 \\
& PointMAE~\cite{pang2022masked}  & 39.18 & 63.13 & 62.09 \\
&PAConv~\cite{xu2021paconv}  & 37.09 & 57.14 & 60.97 \\
\bottomrule
\end{tabular}
\end{center}
\vskip -0.1in
\end{table}

\subsection{Ablation Study}

\textbf{Pre-trained Teacher Models.} Table~\ref{tab: choose teacher} shows the influence of the teacher models on our method's performance. Recent state-of-the-art point cloud networks (PointTransformer~\cite{zhao2021point}, PointNet++~\cite{qi2017pointnet++}, DGCNN~\cite{wang2019dynamic}, PointMAE~\cite{pang2022masked}, and PAConv~\cite{xu2021paconv}) are used as the teacher model. They are all trained on the 3D segmentation task with a large source dataset~\cite{chang2015shapenet}. This table shows that while all recent point cloud networks achieve competitive results, PointNet++~\cite{qi2017pointnet++} shows the best performance. Therefore, we use PointNet++ in all of our experiments.

\begin{figure}[t]
\centering
\subfigure{
\label{fig:rotbot1}
\def\svgwidth{1\columnwidth}
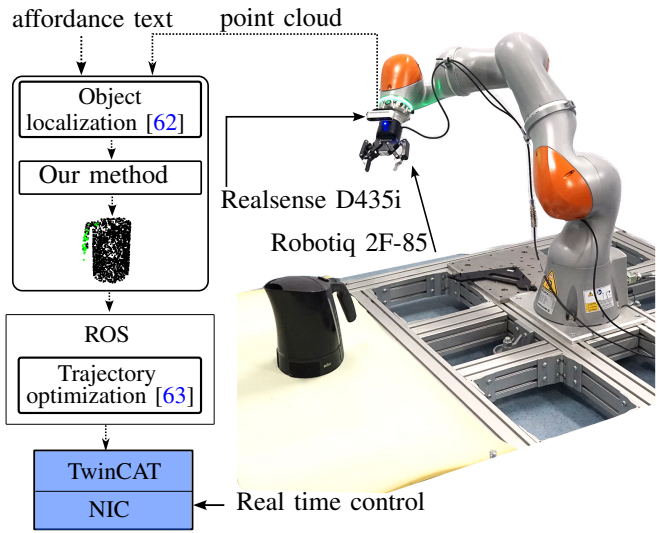
}
%\subfigure{
%\label{fig:robot2}
%\includegraphics[height=20mm]{figures/robot_demonstration/detect_pipeline.pdf}
%}
\vspace{0ex}
\caption{Overview of the robotic experiment.}
\label{fig: robot demonstration}
\end{figure}

%\url{https://drive.google.com/file/d/1XPBdHXrnTiUVcsc9zzdf67GoQcTCve8Z/view?usp=share_li, }
%\textcolor{blue}{done} 
\textbf{Robotic Demonstration.} Figure~\ref{fig: robot demonstration} shows our robotic experimental setup. We used five key components: the \kuka robot, PC1 running Beckhoff TwinCAT software, an Intel RealSense D435i camera, the Robotiq 2F-85 gripper, and PC2 running ROS Noetic 20.04. 
PC1 controls the robot via EtherCAT, while PC2 operates the gripper and camera via USB within ROS. 
These PCs communicate via Ethernet. As in~\cite{ngyen2023open}, we use an object localization method~\cite{zhou2018voxelnet} to identify objects and then sample them to $2048$ points from the scene point cloud. Our framework supports general input commands and generates an affordance region for useful manipulation tasks. The planning and trajectory optimization in~\cite{vu2023machine,beck2022singlularity} is used to execute the action. 
Our Demonstration Video includes several demonstrations that illustrate the versatility of open-vocabulary affordance detection of our method.

\subsection{Discussion}
While our proposed framework has shown significant improvement in open-vocabulary affordance detection in comparison with recent methods, it is imperative to recognize its limitations and potential for future enhancements. Complex semantic affordances, objects with contextual geometry variations, and challenges in novel scenarios can impact our method's effectiveness as we show in the failure cases in Fig.~\ref{fig:fail_case}. Addressing the gap between the semantic concept of text prompts and the geometry of the point cloud is still a challenging problem, especially when the objects' parts share the same geometry but have different affordances. Moving forward, we intend to explore open-vocabulary affordance detection in cluttered scenes to foster quantitative evaluation and direct applications on real robots. Techniques like augmentation, and cross-modal learning~\cite{zhang2023clip} can be useful. Furthermore, combining our open-vocabulary affordance system with long-term manipulation tasks is also an interesting direction~\cite{zhang2018deep}. Finally, as recognized by~\cite{ngyen2023open}, having a new large-scale language-driven affordance dataset with natural point cloud scenes would be more beneficial to real-world robotic applications.

\section{Conclusions}\label{Sec:con}
We have presented a new approach for open-vocabulary affordance detection in 3D point clouds. Our proposed method takes advantage of large-scale pre-trained models and text-point correlation to improve the detection results. By integrating attention mechanisms and knowledge transfer, we outperform other baselines in terms of robustness, generalization, and inference time. These enhancements hold substantial promise to apply our proposed method to different robotic applications. Our source code and trained model will be made publicly available.

%ed OpenAD, a simple yet effective method for open-vocabulary affordance detection in 3D point clouds. Different from traditional approaches, OpenAD, with its capability of semantic understanding, can effectively detect unseen affordances without requiring annotated examples. Empirical results show that OpenAD outperforms other methods by a large margin. We further verified the capability of OpenAD to detect unseen affordances on both known and unseen objects. Additionally, we demonstrated OpenAD's usefulness in real-world robotic applications.  

%While there is much more work to do in this direction, the results of OpenAD are encouraging evidence for intelligent robots comprehend numerous affordances and perform better in complex environments.
%\section*{Acknowledgment}
%\addcontentsline{toc}{section}{Acknowledgment}

\bibliographystyle{class/IEEEtran}
\bibliography{class/IEEEabrv,class/reference}
   
\end{document}